\documentclass[runningheads]{llncs}
\usepackage[T1]{fontenc}
\usepackage{lmodern}
\usepackage{cite}
\usepackage{amsmath,amssymb,amsfonts}
\usepackage[english]{babel}
\usepackage{algorithmic}
\usepackage{graphicx}
\usepackage{subcaption}
\usepackage{xcolor}
\usepackage{hyperref}
\usepackage{multirow}
\usepackage{paralist}
\usepackage{orcidlink}
\usepackage{bbding}

\addto\extrasenglish{%
}
\def\BibTeX{{\rm B\kern-.05em{\sc i\kern-.025em b}\kern-.08em
    T\kern-.1667em\lower.7ex\hbox{E}\kern-.125emX}}

\begin{document}
\title{Compliance-Aware Predictive Process Monitoring: A Neuro-Symbolic Approach}
\titlerunning{Compliance-Aware Predictive Process Monitoring}
%
\author{Fabrizio De Santis\inst{1}\Envelope\orcidlink{0009-0002-5079-3048} \and
Gyunam Park\inst{2}\orcidlink{0000-0001-9394-6513} \and \\
Wil M.P. van der Aalst\inst{3}\orcidlink{0000-0002-0955-6940} \and Francesco Zanichelli\inst{1}\orcidlink{0000-0002-5802-8343}}
\authorrunning{F. De Santis et al.}
%
\institute{University of Parma, Parma, Italy\\\email{\{fabrizio.desantis,francesco.zanichelli\}@unipr.it} \and
Eindhoven University of Technology, Eindhoven, Netherlands\\
\email{g.park@tue.nl} \and
RWTH Aachen University, Aachen, Germany\\
\email{wvdaalst@pads.rwth-aachen.de}}
\maketitle              
\begin{abstract}
Existing approaches for predictive process monitoring are sub-symbolic, meaning that they learn correlations between descriptive features and a target feature fully based on data, e.g., predicting the surgical needs of a patient based on historical events and biometrics.
However, such approaches fail to incorporate domain-specific process constraints (knowledge), e.g., surgery can only be planned if the patient was released more than a week ago, limiting the adherence to compliance and providing less accurate predictions.
In this paper, we present a \textit{neuro-symbolic approach} for predictive process monitoring, leveraging Logic Tensor Networks (LTNs) to inject process knowledge into predictive models.
The proposed approach follows a structured pipeline consisting of four key stages: 1) feature extraction; 2) rule extraction; 3) knowledge base creation; and 4) knowledge injection.
Our evaluation shows that, in addition to learning the process constraints, the neuro-symbolic model also achieves better performance, demonstrating higher compliance and improved accuracy compared to baseline approaches across all compliance-aware experiments.
\keywords{Predictive process monitoring \and Process mining \and Neuro-symbolic AI \and Deep learning \and Reasoning}
\end{abstract}
\section{Introduction}

In business process management, predictive process monitoring has emerged as a vital tool for organizations, enabling them to anticipate process outcomes and irregularities by analyzing historical event data~\cite{DBLP:books/sp/22/FrancescomarinoG22}. 
Current approaches rely heavily on advanced neural networks like LSTMs and Transformers that excel at pattern recognition~\cite{DBLP:conf/caise/TaxVRD17}. 
However, these methods struggle to formally incorporate constraints, domain expertise, and new business rules that may conflict with historical data (e.g., ``surgery can only be planned if a patient was released more than a week ago'').
This prevents purely data-driven models from capturing crucial contextual insights, highlighting the need for hybrid approaches that combine statistical learning with symbolic reasoning.

\textit{Neuro-symbolic artificial intelligence}, which integrates neural networks with symbolic knowledge representation, offers a promising framework to address these challenges~\cite{bhrato_2024}. By embedding domain knowledge, neuro-symbolic approaches enable models to leverage both statistical patterns from data and explicit logical priors.
This paper proposes a neuro-symbolic approach for predictive process monitoring, enhancing neural networks with symbolic reasoning. First, we extract and categorize features from event logs into: \textit{control-flow features}, \textit{temporal features}, and \textit{payload features}. The feature categories define the vocabulary for the rule extraction, with feature types aligning with process rule types (knowledge).
Once extracted, we formalize the rules into a structured knowledge base using first-order logic.

Then, we categorize knowledge types based on their relationships to the classification problem, obtaining three types of process knowledge: \textit{class-dependent knowledge}, which is further categorized into \textit{non-outcome-oriented} and \textit{outcome-oriented} knowledge, and \textit{class-independent knowledge}. 
Finally, this knowledge can be injected into our neuro-symbolic model, leveraging the Logic Tensor Network (LTN) framework~\cite{bagase_2022}. 
The knowledge is injected in three different ways in relation to the main classification task: at the \textit{input level}, by expanding the feature space; at the \textit{output level}, by redefining the classification model's output; and in \textit{parallel}, by incorporating additional knowledge that is not strictly related to the main classification task but can influence the training.

The core idea is to leverage domain knowledge to guide the learning process, ensuring adherence to business rules while improving predictive accuracy. Evaluations on four real-life event logs using a compliance-aware test set show that our approach outperforms traditional deep learning models, such as LSTM and Transformer, and knowledge-encoding techniques, such as semantic loss, in both accuracy and compliance with domain rules, maintaining high constraint adherence even when training data contains few compliant examples.

The paper is structured as follows. In \autoref{sec:soa}, we give an overview of both predictive process monitoring and neuro-symbolic AI fields. 
We introduce preliminary concepts in \autoref{sec:preliminaries}. 
In \autoref{sec:approach}, we show how rules can be mapped to the feature space, what type of rules can be extracted, and how the knowledge base created with these rules can be leveraged by a binary classifier. 
Finally, we assess our approach's capabilities to enhance accuracy and compliance in \autoref{sec:evaluation} and conclude the paper in \autoref{sec:conclusion}.

\section{Related Work}
\label{sec:soa}

Predictive Process Monitoring (PPM) is a branch of process mining that aims to forecast the future states of ongoing business process instances. By analyzing event logs, PPM seeks to predict various aspects, such as the next activities, remaining time, or potential outcomes of processes~\cite{zhou2025process,oyamada2024enhancing}. Deep learning, particularly through Transformer architectures~\cite{DBLP:journals/corr/abs-2104-00721}, LSTM networks~\cite{DBLP:conf/caise/TaxVRD17}, and graph neural models~\cite{amiri2024pgtnet}, has significantly advanced PPM. Although recent work has increasingly focused on integrating fairness principles into predictive process monitoring~\cite{peeperkorn2025achieving}, little attention has been paid to the role of logical constraints and business rules, which remain essential in compliance-sensitive settings. Most existing methods remain predominantly data-driven and ignore constraints that should guide or restrict predictions. Only a small subset of studies incorporates explicit process knowledge: Vazifehdoostirani et al.~\cite{vazifehdoostirani2022encoding} encode control-flow patterns as predefined input features, Di Francescomarino et al.~\cite{di2017eye} enforce LTL constraints as post-processing rules, and Mezini et al.~\cite{DBLP:journals/corr/abs-2509-00834} embed LTL constraints directly into model training through a differentiable loss aimed at improving suffix prediction. However, these strategies have two key limitations: (1) they concentrate mainly on control-flow constraints while paying limited attention to temporal aspects and payload data, and (2) they integrate domain knowledge only at the input or output level (except for~\cite{DBLP:journals/corr/abs-2509-00834}).

Neuro-symblic AI aims to combine the strengths of symbolic reasoning and sub-symbolic learning~\cite{bhrato_2024}. This integration addresses fundamental limitations in both traditional AI paradigms. Approaches like $\delta$ILP~\cite{evgr_2018} and DeepProbLog~\cite{maduki_2018} bring together logic programming paradigms and deep neural models, enabling end-to-end learning by combining neural predicates with probabilistic logic. Other methods, such as semantic loss functions~\cite{DBLP:conf/icml/XuZFLB18} and constraint-aware training methods~\cite{DBLP:conf/icml/FischerBDGZV19}, embed formal rules directly into differentiable objectives. Recent advancements focus on systematic approaches for injecting domain knowledge into neural architectures. Knowledge-Enhanced Neural Networks (KENNs)~\cite{DBLP:conf/pricai/DanieleS19} inject prior knowledge through residual connections, enforcing constraints while maintaining end-to-end learning.

Logic Tensor Networks (LTNs)\cite{bagase_2022} embed first-order logic into neural computational graphs through fuzzy logic semantics, offering a unified paradigm that embeds first-order logic formulas during training. This brings several benefits to the PPM domain: (1) it allows different forms of process knowledge, ranging from control-flow constraints expressed in LTL to payload-dependent business rules, as well as relations involving durations or waiting times; (2) fuzzy semantics enable soft satisfaction of constraints, making them robust to partial compliance, uncertainty, and rule conflicts often present in real process executions; and (3) learning and reasoning become intrinsically coupled, as logical formulas influence the neural optimization process instead of being applied only after model training. Our approach injects knowledge into the reasoning process through a neuro-symbolic architecture that leverages the expressive capabilities of first-order logic. This enables the formulation of interpretable and flexible logical rules. Furthermore, we extend the scope of our analysis to encompass additional aspects of the process, such as data attributes, using new rules and logical constraints that may conflict with the data.

\section{Preliminaries}
\label{sec:preliminaries}

\subsection{Logic Tensor Network (LTN)}
\label{sec:ltn}
Logic Tensor Networks embed first-order logic into differentiable computation graphs, allowing neural models to jointly learn from data and from logical knowledge bases~\cite{bagase_2022}. LTNs ground symbolic objects (constants, functions, predicates) into tensors, evaluate fuzzy connectives, and aggregate the satisfiability of all formulas in a knowledge base. This makes them a convenient substrate for tasks such as classification, relational reasoning, and constraint satisfaction while keeping end-to-end training differentiable.

\begin{figure*}[t]
\centering
\includegraphics[width=0.8\textwidth]{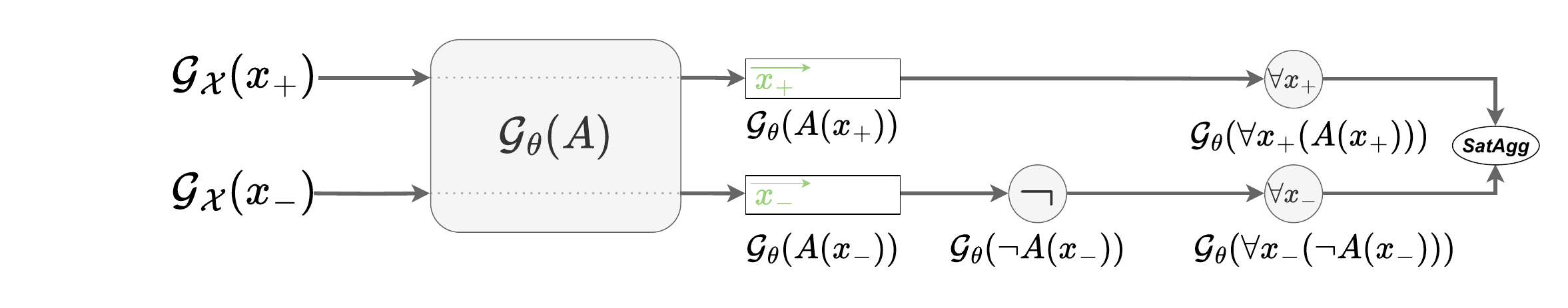}
\caption{Binary classification backbone implemented as an LTN computational graph~\cite{bagase_2022}.} \label{fig:fol-ltn}
\end{figure*}

We build on this general machinery by instantiating an LTN-based binary classifier that serves as the backbone for our compliance-aware extensions. \autoref{fig:fol-ltn} summarizes the resulting computational graph, and the key components are described below.

\paragraph{Grounded inputs.} Each training sample belongs either to the positive set $\mathcal{X}_+$ or the negative set $\mathcal{X}_-$. Variables $x_+$ and $x_-$ range over these two sets. The grounding operator $\mathcal{G}_\mathcal{X}$ maps every variable instantiation to its feature vector, i.e., $\mathcal{G}_\mathcal{X}(x_+)=\{v_1,\dots,v_{|\mathcal{L}_+|}\}$ and analogously for $x_-$. Domain constants (e.g., activity labels, payload values) are grounded via $\mathcal{G}_\mathcal{C}$, whereas deterministic attributes are produced by functions grounded through $\mathcal{G}_\mathcal{F}$.

\paragraph{Predicate evaluation.} The unary predicate $A$ models the binary classifier $\Delta$. Its grounding $\mathcal{G}_\theta(A)$ is a neural network (LSTM or Transformer encoder followed by an MLP) parameterized by $\theta$. Applying $\mathcal{G}_\theta(A)$ to $\mathcal{G}_\mathcal{X}(x_+)$ returns a vector of truth values $\mathcal{G}_\theta(A(x_+))$ with entries in $[0,1]$; values close to~1 indicate samples likely to belong to the positive class. We explicitly model the complementary predicate $\neg A$ by applying the fuzzy negation $1-u$ to the outputs obtained on $\mathcal{X}_-$.

\paragraph{Quantifiers and satisfaction.} Universal quantifiers aggregate truth values for all individuals in $\mathcal{L}_+$ or $\mathcal{L}_-$ via the pMeanError operator:
\begin{equation*}
    \forall x_+\;A(x_+) = 1-\Big(\frac{1}{|\mathcal{L}_+|}\sum_{v\in\mathcal{G}_\mathcal{X}(x_+)}(1-\mathcal{G}_\theta(A(v)))^p\Big)^{\frac{1}{p}},
\end{equation*}
with $p\geq1$ controlling how strictly deviations are penalized (higher $p$ enforces stricter universals). Existential quantifiers rely on the standard generalized mean (pMean). A knowledge base $\mathcal{K}$ aggregates all grounded formulas through $\mathrm{SatAgg}$ (again pMeanError), and the optimization minimizes
\begin{equation*}
    L = 1-\mathrm{SatAgg}_{\phi\in\mathcal{K}}\big(\mathcal{G}_\theta(\phi)\big),
\end{equation*}
thereby tuning $\theta$ so that all constraints jointly approach truth value~1.

This LTN classifier forms the reference pipeline. In Sec.~\ref{sec:approach}, we introduce process-specific logical formulas and show how they extend the backbone by (i) enriching the grounded inputs, (ii) constraining predicate outputs, and (iii) adding parallel objectives that share the same satisfaction operator.

\subsection{Predictive Process Monitoring}
We now instantiate the LTN backbone on process data. An event log records multiple \textit{traces}, each describing one process instance as an ordered sequence of events $\sigma=\langle e_1,\ldots,e_n\rangle$ sharing the same case identifier. Every event $e=(a,c,t,attr_1,\ldots,attr_m)$ captures the executed activity $a$, the case $c$, its timestamp $t$, and optional payload attributes.

Predictive process monitoring targets ongoing instances: a \textit{prefix} $l=(\sigma,k)$ collects the first $k$ events of trace $\sigma$. Let $\mathcal{L}$ be the set of all prefixes extracted from the training logs, with $\mathcal{L}_+$ and $\mathcal{L}_-$ denoting the positive and negative subsets according to the selected outcome label (e.g., complication vs. no complication). The predictive model $\Delta: \mathcal{L}\rightarrow\{+,-\}$ decides whether a new prefix belongs to the positive class.

This setting plugs directly into the notation introduced in \autoref{sec:ltn}: prefixes act as variables $x_+$ and $x_-$, their feature vectors are the groundings $\mathcal{G}_\mathcal{X}$, activity names and payload values populate the constants grounded via $\mathcal{G}_\mathcal{C}$, and deterministic descriptors such as waiting times are produced by functional groundings $\mathcal{G}_\mathcal{F}$. The predicate $A$ becomes the predictive classifier $P$ that scores each prefix, exactly as shown in \autoref{fig:fol-ltn}.

Finally, domain rules can now be written over these symbols. For instance, the constraint \emph{``if antibiotics are not administered within two hours after surgery for an elderly patient, the patient will have complications''} can be expressed as
\[
\forall {l \in \mathcal{L}} \; \big((WaitTime(l,\texttt{Surg},\texttt{ATB})>2\wedge Age(l)>60)\rightarrow P(l)\big)
\]
where $WaitTime, Age \in \mathcal{F}$ and $P$ is the predicate modeling the risk of complications. These formulas will be injected into the LTN knowledge base in Sec.~\ref{sec:approach}.

\section{Neuro-Symbolic AI for Predictive Process Monitoring}
\label{sec:approach}

\begin{figure}[t]
\centering
\includegraphics[width=0.7\columnwidth]{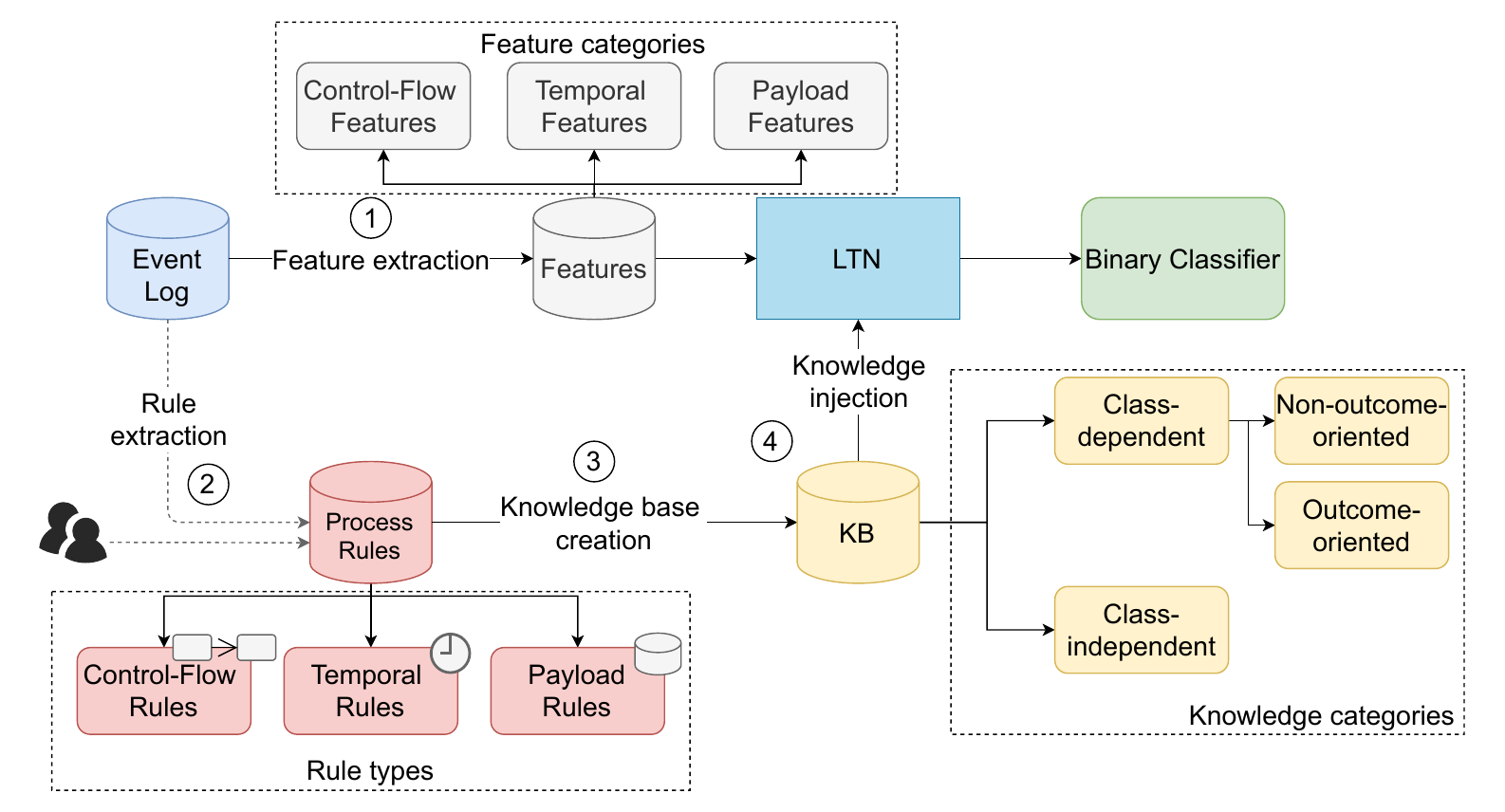}
\caption{The pipeline followed in the approach, which consists of feature extraction, rule extraction, knowledge base creation and injection, and then the creation of the neuro-symbolic model leveraging the LTN framework.} \label{fig:pipeline2}
\end{figure}

Our approach injects process knowledge into predictive process monitoring through the pipeline illustrated in \autoref{fig:pipeline2}. Starting from an event log of traces prefixes, we extract both features and rules, which are then used to build a knowledge base. This knowledge is finally integrated into a binary classifier enriched with process-level constraints.

To illustrate our approach, consider a healthcare scenario where the goal is to predict whether a patient will experience post-surgical complications. A typical case includes preparatory assessments, such as medical history review ($\texttt{Rev}$), physical examinations ($\texttt{Exam}$), laboratory tests ($\texttt{Lab}$), and antibiotic administration ($\texttt{ATB}$), followed by surgery ($\texttt{Surg}$), during which key factors such as procedure duration and intraoperative events are recorded. Postoperative steps include pain management ($\texttt{PAdm}$) and follow-up checks ($\texttt{PostCU}$). In Prefixes $l$ from past executions, together with current and historical patient data, support outcome prediction. We use the predicate $P$ to model the probability that, for a given prefix, the patient will experience post-surgical complications, with $l_+$ and $l_-$ representing positive and negative cases, respectively.

\subsection{Feature Extraction}
\label{sec:featureextraction}
First, we extract features from event logs to capture relevant characteristics of process execution. These features define the vocabulary for both rule extraction and prediction tasks, with the feature space design aligning with the types of process rules (knowledge) we aim to incorporate.

Our approach utilizes three main feature categories:
\begin{compactitem}
    \item \textit{Control-flow features} capture activity sequences based on \textit{Declare}~\cite{DBLP:journals/ife/AalstPS09} constraints. In our healthcare example, these represent the activities performed by the patient and the relationship between them, e.g., $HasAct(l, \texttt{Rev})$ indicates if the activity $\texttt{Rev}$ was performed for a patient.
    \item \textit{Temporal features} capture time-based aspects based on Service Level Agreements (SLAs)~\cite{rafava_2013} on activity durations, waiting times, cycle time, etc. 
    In our healthcare example, this includes the time elapsed between surgery and antibiotic administration, e.g., $WaitTime(l, \texttt{Surg},\texttt{ATB})$.
    \item \textit{Payload features} represent data associated with events and cases, divided into:
    \begin{compactitem}
        \item \textit{Case-level payload}: Data related to the entire case, including numerical, categorical, and unstructured values available at prefix $l$ initiation. In our healthcare example, this includes patient age and pre-existing conditions (diabetes, hypertension, obesity).
        \item \textit{Event-level payload}: Data associated with specific events in the trace prefix $l$, which can only be used once the corresponding activity occurs. In our healthcare example, this includes diagnostic test results, medical reports, or medication administered during treatment.
    \end{compactitem}
\end{compactitem}

\subsection{Rule Extraction}
\label{sec:knowledgeextraction}

The second main step of our pipeline is rule extraction. 
We leverage established process mining techniques to automatically discover and formalize rules from event logs. The types of rules that can be extracted are directly linked to the previously defined feature space, as process rules must map onto appropriate features to be leveraged effectively.

We extract three types of rules:
\begin{compactitem}
    \item \textit{Control-flow rules}: These rules derive from Declarative mining, which identifies constraints such as sequential relationships, mutual exclusion, existence constraints, and choice relationships~\cite{mamova_2011}. The resulting rules can be represented as Linear Temporal Logic (LTL) constraints. In our healthcare example, control-flow rules specify the required order of medical procedures. Using LTL templates like \textit{response}, \textit{chain response}, and \textit{precedence}, we can formalize rules such as:
    \begin{compactitem}
        \item ``If $\texttt{Rev}$ occurs, $\texttt{Exam}$ should occur after $\texttt{Rev}$'' represented as $\square(\texttt{Rev} \implies \lozenge \texttt{Exam})$
        \item ``$\texttt{PostCU}$ should immediately follow $\texttt{Surg}$'' represented as $\square(\texttt{Surg}\implies \bigcirc \texttt{PostCU})$
        \item ``$\texttt{PAdm}$ should occur only if $\texttt{PostCU}$ has occurred before'' represented as $(\neg$$\texttt{PAdm}U$\texttt{PostCU}$)\vee\square(\neg \texttt{PAdm})$
    \end{compactitem}
    \item \textit{Temporal rules}: These rules focus on timing and duration aspects of activities and processes, capturing constraints related to the time elapsed between activities, activity durations, and overall performance. Temporal rules are extracted from SLA compliance analysis~\cite{rafava_2013}, producing \textit{IF-THEN} statements capturing expected timing behaviors. In our healthcare example, a temporal rule might state that ``if $\texttt{ATB}$ happens within two hours after $\texttt{Surg}$, the likelihood of complications decreases.''
    \item \textit{Payload rules}: These rules derive from payload features and capture the data context of process events and cases. We automatically extract payload rules using statistical analysis to uncover correlations between payload attributes and outcomes~\cite{caeppu_2018}. Payload rules are typically expressed in \textit{IF-THEN} format. In our healthcare example, a payload rule might specify that ``if a patient's oxygen saturation falls below 90\% post-surgery, the risk of complications increases.'' 
\end{compactitem}
Domain expertise also plays a crucial role in supplementing automatically extracted rules. For example, knowledge that ``patients with conditions such as diabetes have increased complication risks'' can be manually incorporated into the knowledge base. 

\subsection{Knowledge Base Creation}
\label{sec:kbcreation}
Once extracted, we formalize the rules into a structured knowledge base.

Since our goal is to integrate these rules into an LTN framework, we need to translate the various rule types into first-order logic (FOL) representations.

\subsubsection{Translation of Rules into FOL}
We translate three main types of rules into FOL:
\begin{compactitem}
    \item \textit{Control-flow rules:} We follow the approach in~\cite{DBLP:conf/ijcai/GiacomoV13} to convert LTL rules into FOL, preserving the semantics of temporal operators while allowing for gradient-based optimization. For instance, in a healthcare process example, the constraint $\square(\texttt{Rev} \implies \lozenge \texttt{Exam})$ is translated into $\forall l(HasAct(l, \texttt{Rev}) \wedge Next(l, \texttt{Rev},\texttt{Exam}))$.
    \item \textit{Temporal and payload rules:} We convert \textit{IF-THEN} temporal rules and payload rules into logical formulas by decomposing each rule into its antecedent and consequent. We determine the appropriate quantifier based on the semantics of each antecedent. When a rule should apply universally to all instances, we use the universal quantifier ($\forall$); when it should hold for at least one instance, we use the existential quantifier ($\exists$). For example, the rule ``for patients $l_+$ at risk of complications, if $\texttt{ATB}$ is performed within two hours after $\texttt{Surg}$, the likelihood of complications decreases'' translates to: $\forall l_+(WaitTime(l_+,\texttt{Surg},\texttt{ATB}) \leq 2 \rightarrow \neg P(l_+))$.
\end{compactitem}

\subsubsection{Knowledge Categorization}
We categorize knowledge types based on their relationship to the classification problem:
\begin{compactitem}
    \item \textit{Class-dependent knowledge:} This knowledge type relates to the classes in the classification problem and is further divided into:
    \begin{compactitem}
        \item \textit{Non-outcome-oriented knowledge:} This encodes constraints related to the problem's classes but not directly to the outcome. The main classification predicate does not appear within these constraints. 
            For example, a constraint might establish that ``elderly diabetic patients $l_+$ require special monitoring.''
        
        \item \textit{Outcome-oriented knowledge:} This knowledge directly involves the outcome, including the classification predicate in the consequent of the \textit{IF-THEN} implication. 
            For instance, ``for patients $l_+$ at risk of complications, if $\texttt{ATB}$ is performed within two hours after $\texttt{Surg}$, then the risk of complications decreases.''
    \end{compactitem}
    \item \textit{Class-independent knowledge:} This is general knowledge not tied to the class of the trace, such as control-flow constraints that do not influence the process outcome. 
        For example, a constraint might dictate that ``for patients $l$ $\texttt{Exam}$ should be performed after $\texttt{Rev}$.''
\end{compactitem}

Each knowledge type (control-flow, temporal, and payload) can be classified according to these categories, guiding how they are integrated into the predictive approach.

\subsection{Knowledge Injection}

\begin{figure*}[t]
  \centering
  \includegraphics[width=0.9\textwidth]{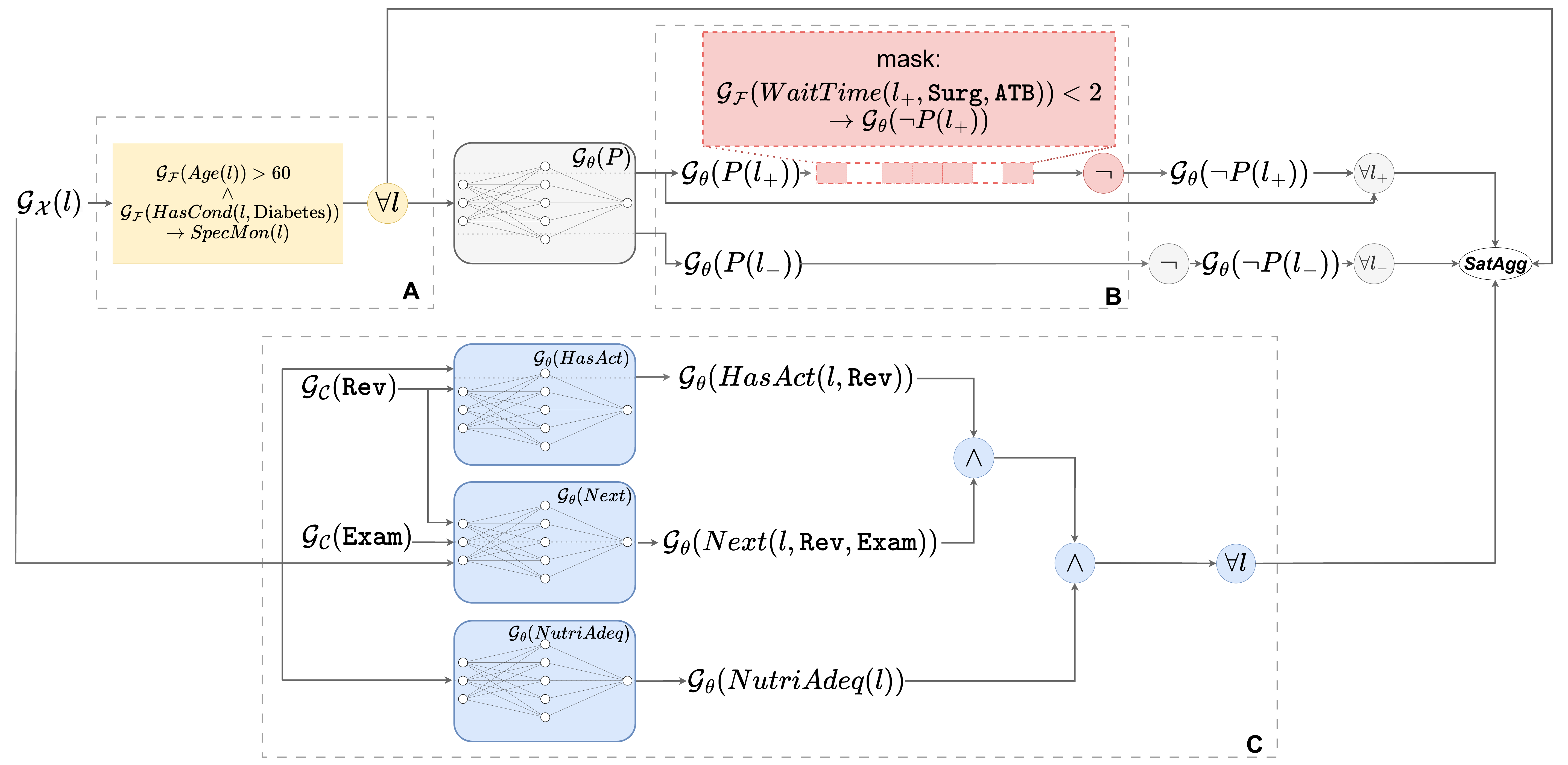}
  \caption{The three ways of injecting knowledge: (A) feature expansion for class-dependent, non-outcome-oriented knowledge, (B) output refinement for class-dependent, outcome-oriented knowledge, and (C) parallel constraints for class-independent knowledge. Each injection pathway addresses a different failure mode of purely data-driven predictors: feature expansion propagates declarative facts such as \emph{``elderly diabetic patients require special monitoring''} into the feature space, output refinement operationalizes outcome rules like \emph{``timely antibiotics lower complication risk''} directly on the predicate outputs, and parallel constraints regularize the shared representation by penalizing process executions structural rules such as \emph{``medical history review must precede physical examination.''}} \label{fig:approach}
  \end{figure*}

After creating the knowledge base, we inject this knowledge into our neuro-symbolic model. 
We propose three distinct injection methods based on the three knowledge categories.
As shown in \autoref{fig:approach}, each method injects knowledge at different points relative to the binary classification predicate \textit{P}, which serves as the core component of our prediction system.

\subsubsection{Feature Expansion}

This method injects \textit{class-independent, non-outcome-oriented knowledge} by preprocessing data before it reaches the binary classification predicate \textit{P} (block A in \autoref{fig:approach}, highlighted in yellow).
This knowledge enhances the feature space with additional information without directly referencing the classification outcome.

Since these constraints serve to expand the feature space, they are implemented as predicates modeled using functions that return a deterministic value in the interval $[0,1]$ when evaluated on a trace prefix. To ensure their integration within the overall reasoning process, these logical rules are also incorporated into the knowledge base, allowing them to interact with other knowledge and contribute directly to the optimization objective by influencing the loss function.
We apply constraints universally to all instances using the $\forall$ operator, then pass the augmented representations to the classification predicate \textit{P}.

In our healthcare scenario, the knowledge that ``elderly diabetic patients require special monitoring'' can be expressed as:
\[
Age(l) > 60 \wedge HasCond(l, \text{Diabetes}) \rightarrow SpecMon(l)
\]
The premise of this formula is used to generate a new feature indicating whether both conditions are satisfied for each patient $l$, where $Age$ returns the patient's age and $HasCond$ returns $1$ if the patient has diabetes and $0$ otherwise. The values produced by evaluating such logical rules are then concatenated with the original feature vector representation.

\subsubsection{Output Refinement}

This method injects \textit{class-dependent, outcome-oriented knowledge} by modifying the outputs of the classification predicate (block B in \autoref{fig:approach}, highlighted in red). 
These constraints directly influence prediction outcomes by imposing rules on the outputs $P(l_+)$ and $P(l_-)$.
This configuration has the most direct impact on final predictions since it refines the classification results rather than modifying the input feature space. 
The constraints can adjust prediction probabilities based on domain rules that explicitly reference the outcome.
In our healthcare scenario, the rule ``timely antibiotic administration reduces complication risk'' can directly modify the complication probability estimated by predicate \textit{P}. 
When antibiotics are administered within two hours of surgery, the constraint decreases the truth value for positive prefixes ($P(l_+)$).

\subsubsection{Parallel Constraints}

This method injects \textit{class-independent knowledge} as parallel constraints that operate alongside the main classification task (block C in \autoref{fig:approach}, highlighted in blue).
Since this knowledge does not depend on outcomes, it applies to all prefixes $l$.
These constraints are implemented as auxiliary predicates modeled by separate neural networks, serving as complementary objectives that must be satisfied jointly with the main classification task.

Although they do not reference the class label, their gradients regularize the shared representation by penalizing process executions that violate structural rules. This prevents the classifier from overfitting to non-compliant patterns that accidentally correlate with the outcome.
To ensure the optimization of the primary predictor is not adversely influenced, only prefixes that conform to the encoded domain rules are incorporated into the parallel constraint-related loss. Prefixes that violate these rules are therefore excluded from this loss component, preventing non-conformant behavior from introducing misleading gradients during training.
The truth values of these auxiliary predicates are aggregated and connected to the main satisfaction operator so that satisfying the structural rules increases the overall SatAgg score.

In our healthcare scenario, process-structural knowledge such as ``medical history review must precede physical examination'' and ``patients should follow proper nutrition guidelines'' can be encoded as
\[
\forall l\, \Big((HasAct(l, \texttt{Rev}) \wedge Next(l, \texttt{Rev}, \texttt{Exam})) \wedge NutriAdeq(l)\Big)
\]
where $HasAct$ is a predicate modeling the probability that $\texttt{Rev}$ happened in $l$, $Next$ is the predicate modeling the probability that $\texttt{Exam}$ follows $\texttt{Rev}$, and $NutriAdeq$ is a predicate modeling the probability that the patient is following a proper diet. Enforcing these constraints encourages the classifier to rely on medically plausible prefixes; when such structure correlates with the outcome (e.g., well-structured care often yields fewer complications), indirect performance gains are observed.

\section{Evaluation}
\label{sec:evaluation}

\subsection{Datasets}

We evaluated our approach on an outcome prediction task using four publicly available real-life event logs. 
We selected the event logs based on the presence of case-level and event-level attributes that could serve as domain knowledge to define meaningful logical rules and influence outcome prediction.
\begin{compactitem}
    \item \textit{BPIC2012}: This event log pertains to the loan application process of a Dutch bank. We defined the labeling based on whether an application is accepted or not.
    \item \textit{BPIC2017}: This is a higher quality version of BPIC2012 with more examples and features. The labeling is based on whether an application is accepted or not.
    \item \textit{Traffic fines}: A real-life event log of a system managing traffic fines. We defined the labeling based on whether the fine is sent for credit collections or paid in full.
    \item \textit{Sepsis}: This event log contains sepsis cases from a hospital. We defined the outcome label based on whether the patient is admitted to the ICU or not.
\end{compactitem}

\subsection{Experimental Design}

In our experiments\footnote{The code to reproduce our experiments is available at \url{https://github.com/FabrizioDeSantis/NeSyPPM-Compliance}.}, we compare eight architectural variants to evaluate the impact of incorporating domain knowledge into predictive process monitoring:
\begin{compactenum}
    \item \textbf{LSTM}~\cite{DBLP:conf/caise/TaxVRD17}, \textbf{TFR}~\cite{DBLP:journals/corr/abs-2104-00721}: Purely data-driven baselines trained with binary-cross entropy. 
    \item \textbf{LSTM-FE}, \textbf{TFR-FE}: LSTM and Transformer models with knowledge encoded in the feature space.
    \item \textbf{LSTM-SL}, \textbf{TFR-SL}: LSTM and Transformer models trained with semantic loss~\cite{DBLP:conf/icml/XuZFLB18}.
    \item \textbf{LTN-Data-L}, \textbf{LTN-Data-T}: LTN~\cite{bagase_2022} with LSTM/Transformer backbones trained without additional domain knowledge.
    \item \textbf{LTN-L}, \textbf{LTN-T}: Our proposed approach incorporates domain-specific process knowledge through the neuro-symbolic framework (i.e., configurations A, B, and C).
\end{compactenum}

The LSTM model has two layers and a hidden dimension of 128. The transformer architecture uses a hidden size of 128, along with two attention heads. Both models are trained using the Adam optimizer with a learning rate of 0.001 and a batch size of 32. Categorical variables are represented through 64-dimensional embeddings, while numerical features are normalized and then processed using linear projection layers. All models are trained for 100 epochs, with early stopping applied based on validation set performance to prevent overfitting. While more complex neural architectures could be considered, we intentionally used a simplified design to isolate and clearly evaluate the impact of injected knowledge constraints on prediction performance. Our approach is architecture-agnostic, with results expected to transfer to other neural network designs. We do not compare with knowledge-enhanced methods for next suffix prediction, as these techniques encode only control-flow information and are structurally tied to sequence continuation. Adapting them to outcome prediction would require redesigning their core modeling assumptions and would not provide a meaningful and fair baseline. Similarly, we do not consider post-hoc output refinement with hard rule enforcement, which acts as an oracle by imposing constraints after prediction rather than shaping the learned representations. In contrast, our framework integrates process knowledge directly during training, enabling the model to learn under uncertainty and noise rather than relying on external corrections.

For evaluation, we adopted an 80-20 train-test split, reserving 20\% of the training as a validation set. The test set consists of 1) \textit{rule-compliant traces} that satisfy both their original outcome label and the consequent of applicable rules and 2) randomly selected examples from the original dataset.

This design specifically evaluates how well models adapt to new compliance constraints that may not be represented in historical training data. 
Note that the knowledge base associated with each event log consists of six logical rules. 

Our evaluation focuses on three key aspects:
\begin{compactenum}
    \item \textbf{RQ1}: How does the injection of domain knowledge improve predictive performance compared to purely data-driven approaches?
    \item \textbf{RQ2}: How do the different knowledge injection configurations in the enhanced LTN (configurations A, B, and C in \autoref{fig:approach}) influence predictive performance?
    \item \textbf{RQ3}: How well do the models adhere to domain constraints, as measured by the proportion of satisfied constraints over all applicable constraints? A constraint is considered applicable when a prefix satisfies the constraint's conditions, and it is satisfied when the model's prediction complies with the constraint.
\end{compactenum}

\subsection{Experimental Results}
Our evaluation demonstrates significant performance differences among the three architectural variants. 
As shown in \autoref{tab:evaluation}, the enhanced LTN with domain knowledge consistently outperforms both the baseline LSTM and basic LTN models.

\subsubsection{RQ1: Prediction Performance}

\begin{table}[!t]
\centering
\caption{Performance on compliance-aware test set comparing purely data-driven baselines (LSTM, TFR, LTSM-FE, TFR-FE) with knowledge-enhanced variants. Results show mean and standard deviation over 5 seeds.}
\label{tab:evaluation}
\resizebox{\columnwidth}{!}{%
\begin{tabular}{ccccccccc}
\hline
\multirow{2}{*}{\textbf{Model}} &
  \multicolumn{2}{c}{\textbf{Sepsis}} &
  \multicolumn{2}{c}{\textbf{BPIC2012}} &
  \multicolumn{2}{c}{\textbf{BPIC2017}} &
  \multicolumn{2}{c}{\textbf{Traffic fines}} \\ \cline{2-9} 
                    & Accuracy     & F1           & Accuracy     & F1                    & Accuracy     & F1           & Accuracy     & F1           \\ \hline
\textbf{LSTM}       & 81,86 ± 1,57 & 71.61 ± 4.70 & 53.19 ± 2.45 & 52.49 ± 1.99          & 64.62 ± 0.76 & 64.04 ± 1.15 & 77.34 ± 1.54 & 77.27 ± 1.65 \\
\textbf{TRF}        & 81,53 ± 2,31 & 73.66 ± 2.78 & 58.80 ± 1.21 & 56.15 ± 1.09          & 69.22 ± 3.11 & 68.93 ± 3.53 & 78.26 ± 1.51 & 78.23 ± 1.52 \\
\textbf{LSTM-FE}    & 81,71 ± 1.60 & 70.49 ± 3.87 & 51.89 ± 0.97 & 51.80 ± 1.03          & 63.35 ± 0.54 & 62.20 ± 0.49 & 77.15 ± 0.61 & 77.05 ± 0.69 \\
\textbf{TRF-FE}     & 78.61 ± 3.34 & 67.01 ± 4.09 & 62.29 ± 2.49 & 58.83 ± 2.16          & 66.41 ± 1.64 & 65.77 ± 1.97 & 75.18 ± 0.81 & 74.90 ± 0.92 \\
\textbf{LSTM-SL}    & 79.90 ± 2.72 & 69.70 ± 4.59 & 53.70 ± 1.94 & 53.08 ± 1.52          & 64.98 ± 0.59 & 64.34 ± 0.63 & 77.54 ± 1.42 & 77.46 ± 1.52 \\
\textbf{TRF-SL}     & 78.60 ± 1.25 & 67.66 ± 2.35 & 64.24 ± 3.27 & 60.98 ± 1.95          & 68.74 ± 1.70 & 68.28 ± 2.08 & 78.88 ± 0.25 & 78.83 ± 0.25 \\
\textbf{LTN-Data-L} & 81.44 ± 2.50 & 71.31 ± 5.37 & 54,60 ± 0,89 & 53,61 ± 0,92          & 71.33 ± 0.14 & 70.85 ± 0.20 & 76.65 ± 0.66 & 76.53 ± 0.70 \\
\textbf{LTN-Data-T} & 80.31 ± 1.61 & 74.54 ± 3.33 & 60,33 ± 3,81 & 57,51 ± 2,84          & 75.07 ± 0.81 & 74.21 ± 0.48 & 77.09 ± 1.12 & 76.94 ± 1.27 \\
\textbf{LTN-L}      & 92.68 ± 0.75 & 91.18 ± 0.56 & 64.50 ± 1.76 & \textbf{62.59 ± 0.49} & 76.59 ± 0.14 & 75.10 ± 0.33 & 79.89 ± 0.50 & 79.89 ± 0.50 \\
\textbf{LTN-T} &
  \textbf{93.15 ± 1.59} &
  \textbf{92.06 ± 1.59} &
  \textbf{64.66 ± 3.44} &
  61.41 ± 2.20 &
  \textbf{76.99 ± 0.37} &
  \textbf{75.40 ± 0.31} &
  \textbf{80.20 ± 0.25} &
  \textbf{80.19 ± 0.25} \\ \hline
\end{tabular}%
}
\end{table}

LTNs enriched with domain knowledge outperform both purely data-driven counterparts and models using semantic loss across all datasets. The performance gains are more pronounced on smaller datasets: \textit{Sepsis} achieves an F1 improvement of 18.4\% (TFR$\rightarrow$LTN-T), and \textit{BPIC12} improves by 10.1\% (LSTM$\rightarrow$LTN-L). On larger datasets, such as \textit{Traffic fines}, the improvements are more modest (1.96\% TFR$\rightarrow$LTN-T). This attenuation is likely due to the abundance of training samples, which reduces the relative influence of logical rules during learning. Notably, the model achieves performance improvements even in settings where the number of compliant traces in the training set is limited (3.18\% for \textit{Sepsis} and 13.09\% for \textit{BPIC12}). Overall, these results highlight the effectiveness of incorporating domain knowledge into predictive process monitoring, especially when data is limited.

\begin{figure*}[!h]
  \centering
  \begin{subfigure}[b]{0.33\textwidth}
    \includegraphics[width=\textwidth]{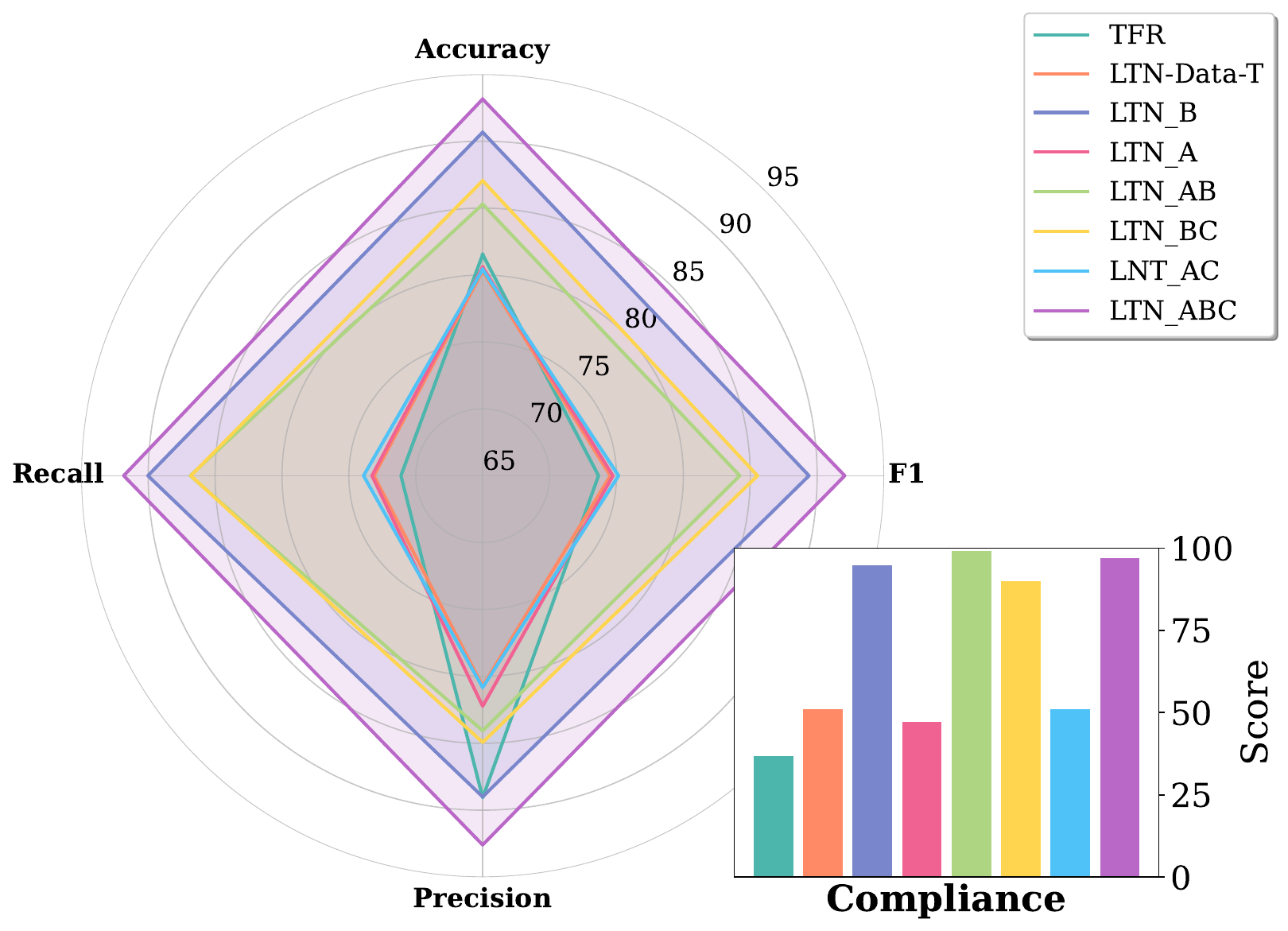}
    \caption{Sepsis}
    \label{fig:sepsis}
  \end{subfigure}
  \begin{subfigure}[b]{0.33\textwidth}
    \includegraphics[width=\textwidth]{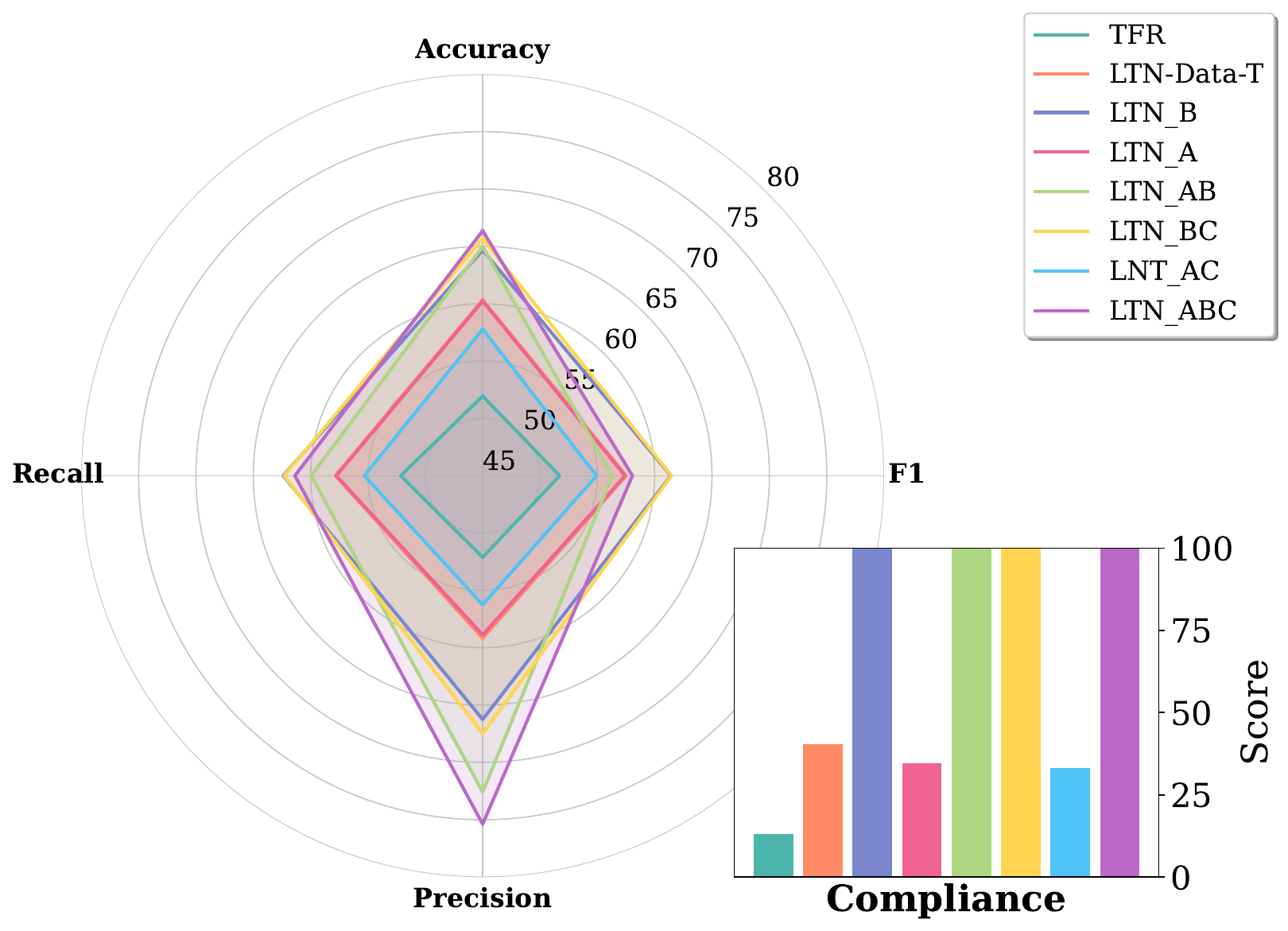}
    \caption{BPIC2012}
    \label{fig:bpi12}
  \end{subfigure}
  \begin{subfigure}[b]{0.33\textwidth}
    \includegraphics[width=\textwidth]{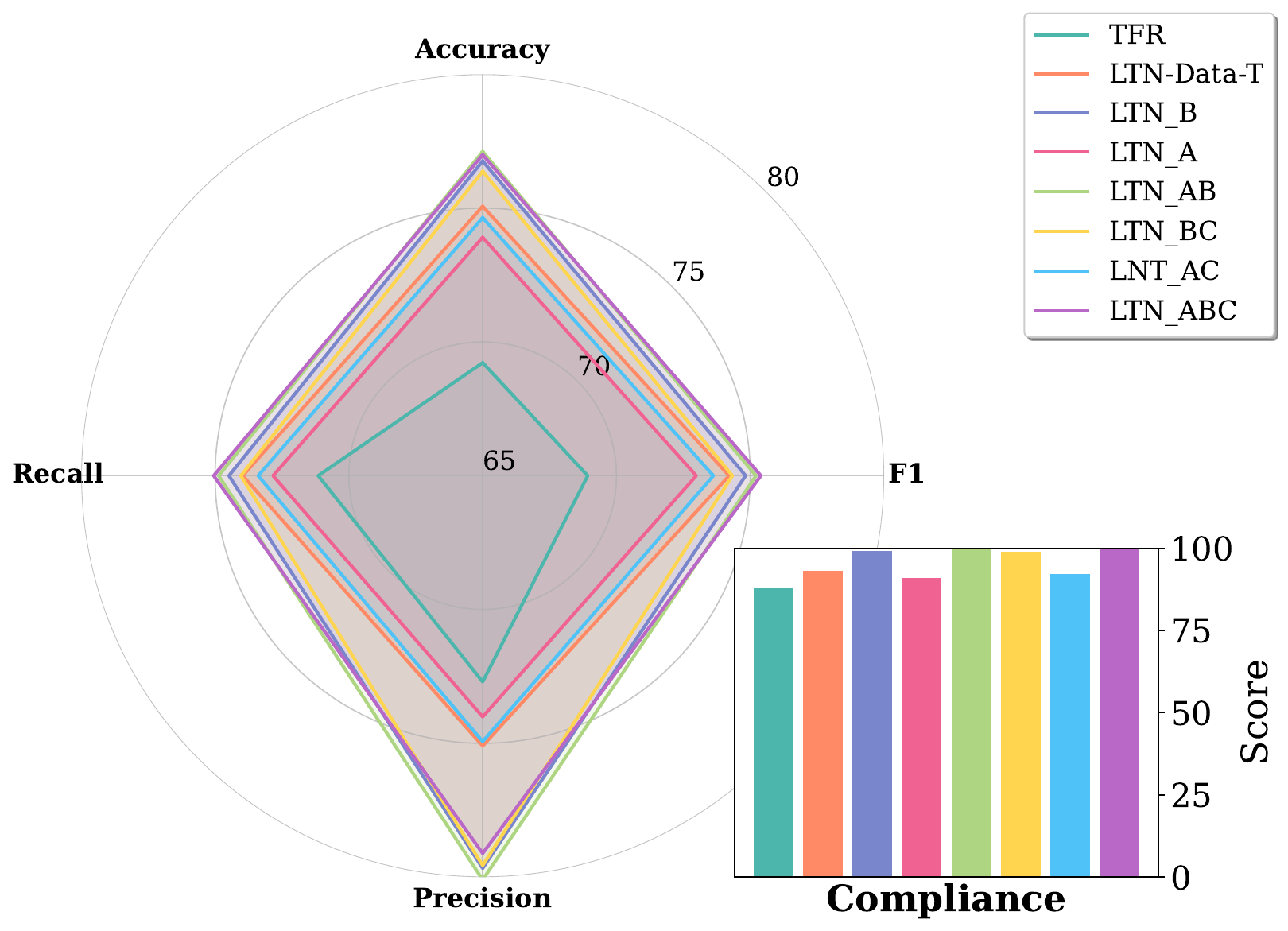}
    \caption{BPIC2017}
    \label{fig:bpi17}
  \end{subfigure}
  \begin{subfigure}[b]{0.33\textwidth}
    \includegraphics[width=\textwidth]{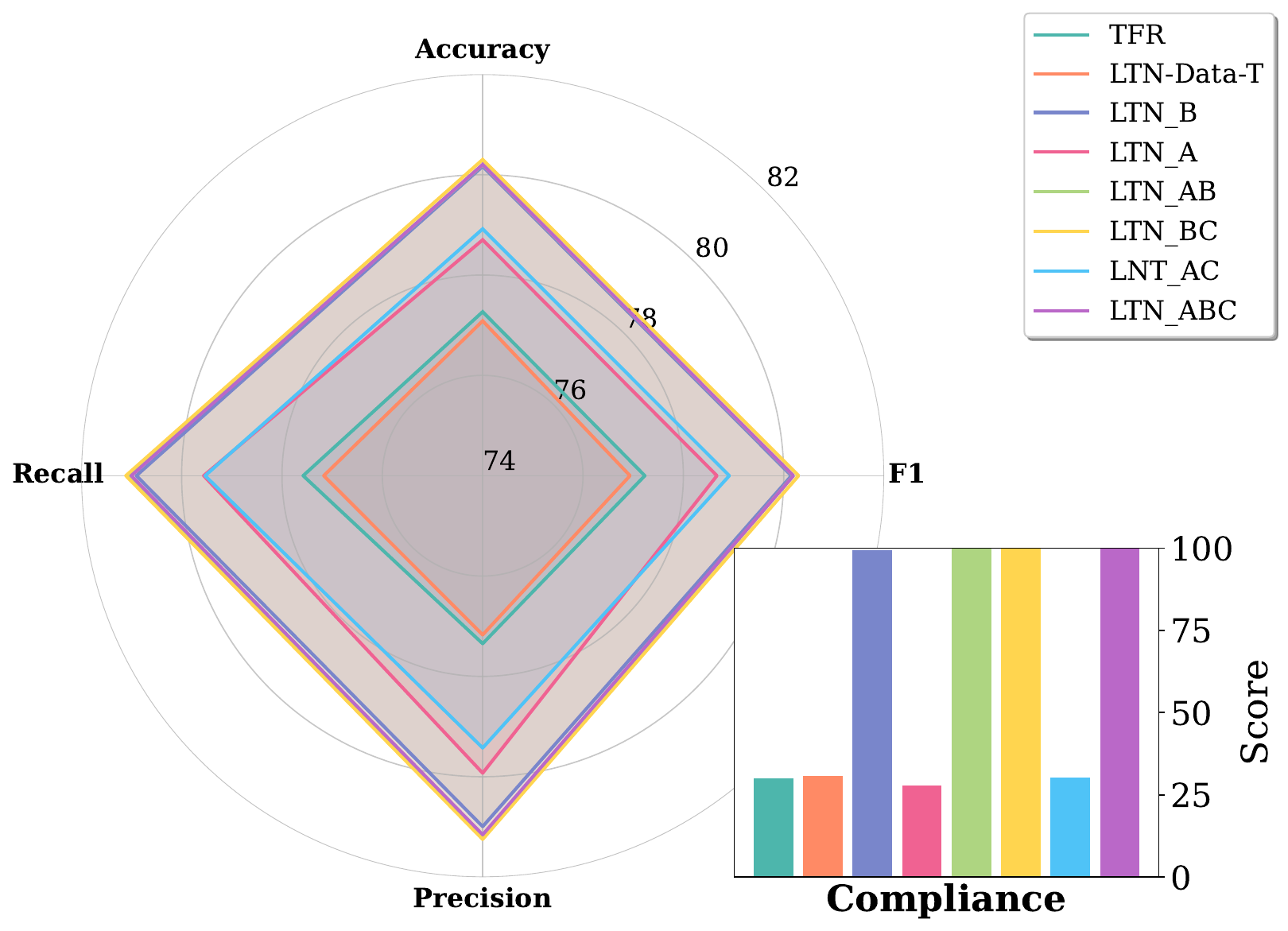}
    \caption{Traffic fines}
    \label{fig:traffic}
  \end{subfigure}
  \caption{Ablation study for LTN-T of the different injection methods with 5 different seeds. Injection techniques were first evaluated individually (i.e., feature expansion LTN\_A and output refinement LTN\_B), followed by an evaluation of their possible combinations, where the LTN\_ABC is the main configuration. The letters used to denote the configurations correspond to the injection methods in ~\autoref{fig:approach}.}
  \label{fig:evaluation}
\end{figure*}

\subsubsection{RQ2: Knowledge Injection Impact}
We analyzed the impact of different neuro-symbolic integration strategies. The results highlight architectural trade-offs (see \autoref{fig:evaluation}):
\begin{compactitem}
    \item Output Refinement (B): Consistently delivered the strongest performance improvements, effectively leveraging new compliance constraints even when training data contained few compliant examples.
    \item Feature Expansion (A): Demonstrated limited effectiveness compared to output refinement, as reflected in both classification metrics and compliance scores.
    \item Parallel Constraints (C): Adding constraints unrelated to the main prediction task slightly improves classification performance when combined with configuration B. While such constraints typically increase computational complexity when modeled using neural networks, they are implemented as deterministic functions without additional neural models. These constraints might prove valuable in multi-task scenarios, such as simultaneously predicting the next activity.
\end{compactitem}

\begin{figure*}[!t]
  \centering
  \begin{subfigure}[b]{0.45\textwidth}
    \includegraphics[width=\textwidth]{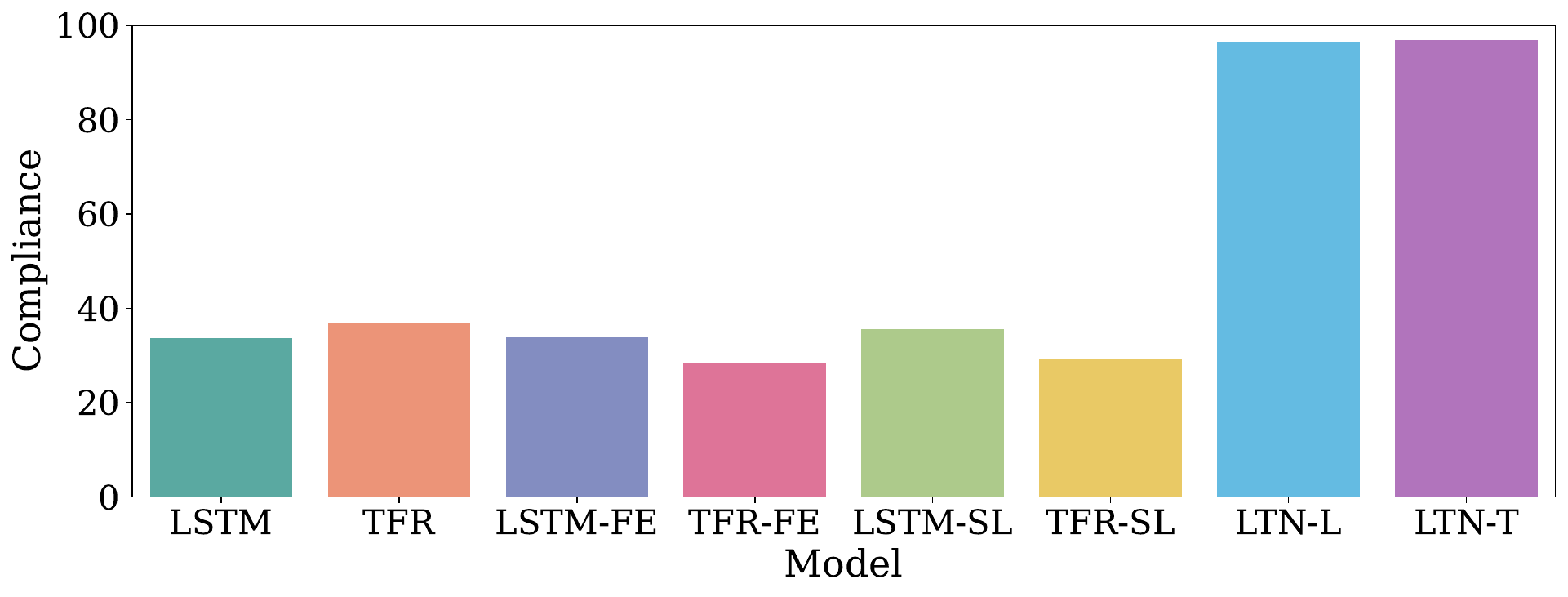}
    \caption{Sepsis}
    \label{fig:sepsis_comp}
  \end{subfigure}
  \begin{subfigure}[b]{0.45\textwidth}
    \includegraphics[width=\textwidth]{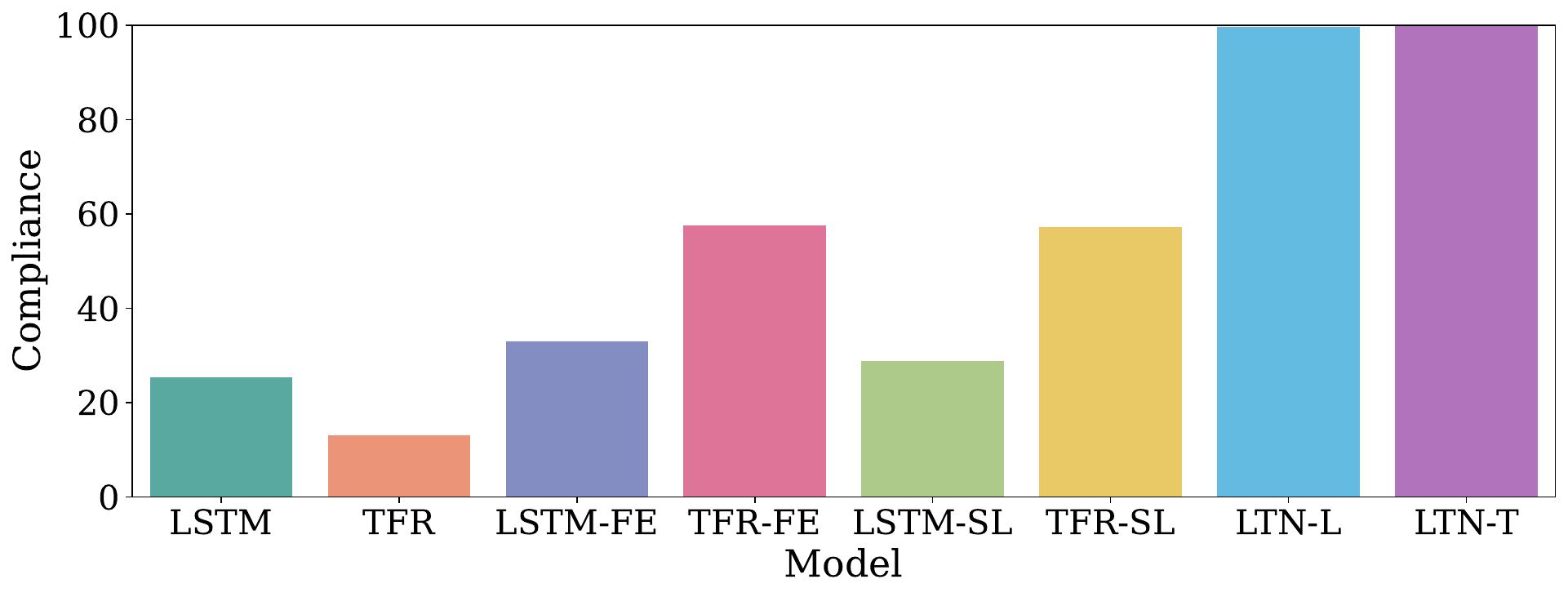}
    \caption{BPIC2012}
    \label{fig:bpi12_comp}
  \end{subfigure}
  \begin{subfigure}[b]{0.45\textwidth}
    \includegraphics[width=\textwidth]{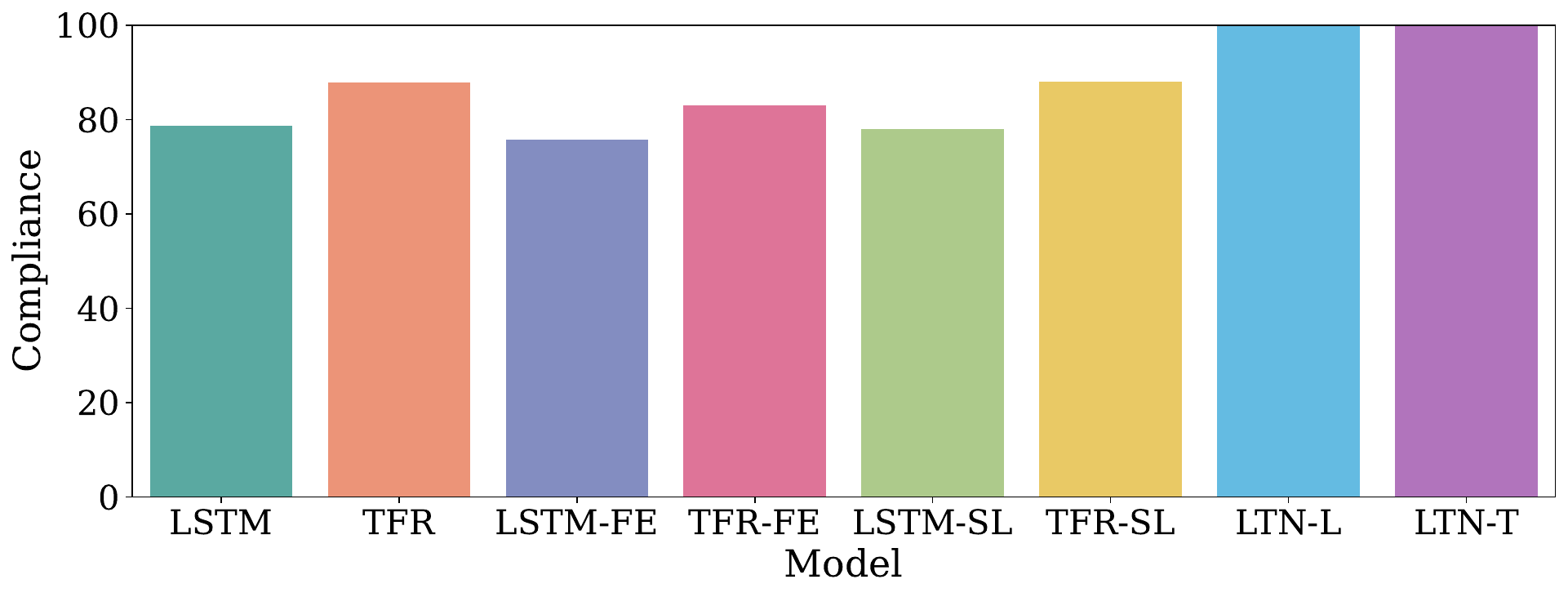}
    \caption{BPIC2017}
    \label{fig:bpi17_comp}
  \end{subfigure}
  \begin{subfigure}[b]{0.45\textwidth}
    \includegraphics[width=\textwidth]{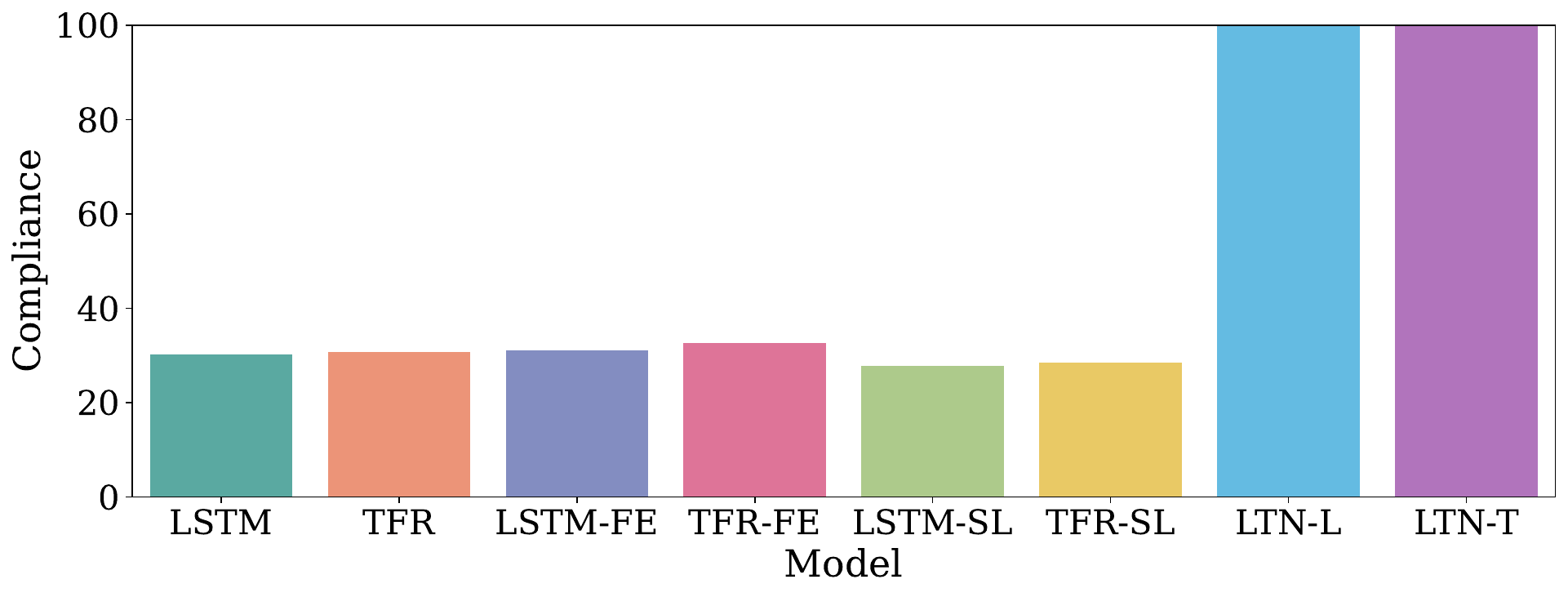}
    \caption{Traffic fines}
    \label{fig:traffic_comp}
  \end{subfigure}
  \caption{Compliance scores obtained with 5 different seeds comparing purely data-driven baselines with knowledge-enhanced variants.}
  \label{fig:evaluation_compliance}
\end{figure*}

\subsubsection{RQ3: Compliance Assessment}
A key advantage of the enhanced LTN is its ability to consistently satisfy the imposed logical constraints by directly encoding them into the training process, achieving high compliance scores even in settings where compliant traces are scarce. While semantic-loss models may in some cases yield improvements over purely data-driven approaches, such gains are not consistent, as the semantic-loss functions act merely as regularizers. As shown in \autoref{fig:evaluation_compliance}, the baseline models exhibit competitive compliance scores on the \textit{BPIC2017} log, largely due to the high prevalence of compliant traces in the training set (43.92\%), which allows statistical correlations to capture valid patterns.
However, purely data-driven approaches struggle to generalize when introduced to new logical constraints or when compliant patterns are statistically rare, as observed in \textit{Sepsis} (4.55\%) and \textit{BPIC2012} (13.09\%). In contrast, our neuro-symbolic approach explicitly embeds these constraints into the learning objective. Consequently, the enhanced LTN demonstrates superior adaptability, maintaining high compliance scores even when the historical training distributions do not heavily support specific process rules.

\section{Conclusion}
\label{sec:conclusion}

In this work, we presented a neuro-symbolic approach based on LTNs for predictive process monitoring, leveraging business rules and logical constraints to enhance predictive and compliance adherence in compliance-aware settings. Experimental results showed that our method can effectively leverage injected constraints. Our approach demonstrated the ability to build a reliable predictive model, thanks to the injected knowledge, even when the training set lacks compliant traces. These findings show the potential of neuro-symbolic learning in predictive process monitoring, especially in scenarios where domain knowledge and logical consistency are critical. In future work, we plan to formalize an explicit representation defining specific templates to automatically connect features with rules. Moreover, we aim to evaluate our neuro-symbolic approach on a case study where domain knowledge plays a critical role, such as in healthcare. Finally, we plan to extend the approach to more process-aware tasks, such as predictions related to next events and time-related aspects.

\section*{Acknowledgements}

F. De Santis is supported by the Italian Ministry of University and Research (MUR) under the National Recovery and Resilience Plan (NRRP), Mission 4, Component 1, Investment 4.1, CUP D91I23000080006, funded by the European Union - NextGenerationEU.

%
%
%
\bibliographystyle{splncs04}
\bibliography{mybibliography}
\end{document}